%% file: main.tex
\documentclass[10pt]{article}
\usepackage[preprint]{tmlr}
\usepackage{stfloats}
\usepackage{enumitem}
\input{math_commands.tex}

\definecolor{kleinblue}{rgb}{0.0,0.184,0.655}

\usepackage{hyperref}
\usepackage{xcolor}
\hypersetup{
    colorlinks=true,
    linkcolor=kleinblue,
    citecolor=kleinblue
}

\usepackage{url}
\usepackage{booktabs}
\usepackage{threeparttable}
\usepackage{caption}
\usepackage{multirow}
\usepackage{float}
\usepackage{array}
\usepackage{amsmath}
\usepackage{graphicx}
\usepackage{tablefootnote}
\usepackage{siunitx}
\title{\textcolor{kleinblue}{AraLingBench}\\ A Human-Annotated Benchmark for Evaluating \\Arabic Linguistic Capabilities of Large Language Models}
\author{\name Mohamad Zbib\thanks{\textbf{Equal contribution.} \\ This work was done during Mohammad's internship at KAUST.} \\
\addr King Abdullah University of Science and Technology (KAUST) \\
American University of Beirut (AUB) \\
\email mohamad.zbib@kaust.edu.sa
\AND
\name Hasan Abed Al Kader Hammoud\footnotemark[1] \\ 
\addr King Abdullah University of Science and Technology (KAUST) \\
\email hasanabedalkader.hammoud@kaust.edu.sa
\AND
\name Sina Mukalled \\
\addr American University of Beirut (AUB) \\
\AND
\name Nadine Rizk \\
\addr American University of Beirut (AUB) \\
\AND 
\name Fatima Karnib \\
\addr American University of Beirut (AUB) \\ 
\AND
\name Issam Lakkis \\
\addr American University of Beirut (AUB) \\
\AND 
\name Ammar Mohanna \\
\addr American University of Beirut (AUB) \\
\AND
\name Bernard Ghanem \\
\addr King Abdullah University of Science and Technology (KAUST)
}

\def\month{MM}
\def\year{YYYY}
\def\openreview{\url{https://openreview.net/forum?id=XXXX}}

\begin{document}

\maketitle

\begin{abstract}
We present \textit{AraLingBench}, a fully human annotated benchmark for evaluating the Arabic linguistic competence of large language models (LLMs). The benchmark spans five core categories: grammar, morphology, spelling, reading comprehension, and syntax, through 150 expert-designed multiple choice questions that directly assess structural language understanding. Evaluating 35 Arabic and bilingual LLMs reveals that current models demonstrate strong surface level proficiency but struggle with deeper grammatical and syntactic reasoning. AraLingBench highlights a persistent gap between high scores on knowledge-based benchmarks and true linguistic mastery, showing that many models succeed through memorization or pattern recognition rather than authentic comprehension. By isolating and measuring fundamental linguistic skills, \textit{AraLingBench} provides a diagnostic framework for developing Arabic LLMs. The full evaluation code is publicly available on \href{https://github.com/hammoudhasan/AraLingBench}{GitHub}.

\end{abstract}

\begin{figure}[H]
    \centering
    \includegraphics[width=1\linewidth]{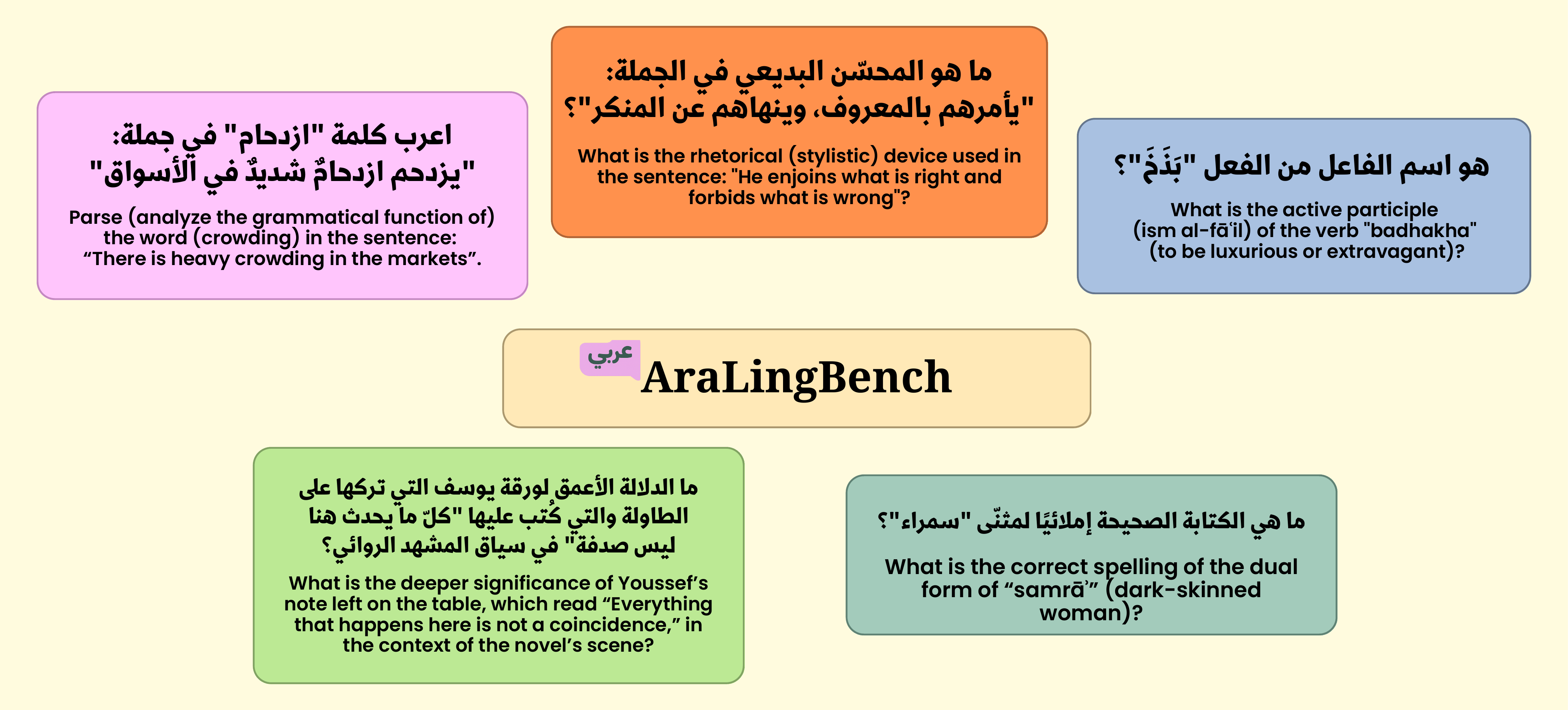}
\caption{\textbf{Sample Questions from AraLingBench.} 
Example items illustrating the five linguistic categories: grammar, morphology, spelling, reading comprehension, and syntax. Each question targets a distinct aspect of Arabic linguistic competence and is crafted by expert annotators to assess genuine linguistic understanding.}    \label{fig:samples}
\end{figure}

\section{Introduction}
\label{sec:Introduction}

Natural language processing (NLP) in Arabic has seen remarkable progress in recent years, supported by the growing availability of large language models (LLMs) trained specifically for Arabic or Arabic-English bilingual contexts. These models range from general-purpose systems to domain-focused variants \citep{al-khalifa202501340v2, inoue-etal-2021-interplay}, and collectively have advanced Arabic NLP in tasks of generation, reasoning, and understanding. However, this rapid growth in model development has not been accompanied by an equally systematic approach to evaluation. The field still lacks reliable methods to assess how well these models truly \textit{understand} Arabic at the linguistic level.

Existing benchmarks such as BALSAM \citep{sengupta2024balsam}, CamelEval \citep{qian2024camelevaladvancingculturallyaligned}, and 3LM \citep{boussaha-etal-2025-3lm} offer a broad coverage of reasoning, comprehension, and task-specific evaluation. However, they share a common limitation: their strong emphasis on factual recall and problem-solving often comes at the expense of testing fundamental language competence. Datasets like EXAMS and ArabicMMLU primarily assess academic or general knowledge, rarely examining whether models can correctly handle grammatical agreement, morphological derivation, or orthographic conventions. Even recent efforts that incorporate dialectal content, such as DialectalArabicMMLU \citep{altakrori202527543v1}, remain largely within this knowledge-oriented paradigm. Consequently, the community lacks a benchmark that isolates and measures the linguistic foundations underlying genuine Arabic language understanding.

Arabic presents unique linguistic challenges that differ substantially from those in English and other Indo-European languages. It is characterized by complex morphology, a rich system of inflection and derivation, and flexible syntax that allows multiple valid word orders. Thus, mastery of Arabic requires competence across several interrelated skills presented in Figure  \ref{sec:Introduction}: grammar (\textit{Nahw}), morphology (\textit{Sarf}), orthography (\textit{Imlaa}), reading comprehension (\textit{Fahm al-logha}), and syntactic structure (\textit{Tarkib Lughawi}). However, despite their central importance to the language, these aspects are often assumed rather than directly tested in current evaluation practices.

To address this gap, we introduce \textbf{AraLingBench}, a human-annotated benchmark specifically designed to evaluate Arabic LLMs on their core linguistic competence. The benchmark consists of 150 multiple-choice questions authored by experts, distributed evenly across the five categories above. Each question is written and reviewed by trained Arabic linguists to ensure linguistic validity, pedagogical clarity, and a focus on reasoning grounded in the language itself rather than factual recall. Figure~\ref{fig:samples} shows representative examples of these categories, illustrating how AraLingBench isolates different dimensions of Arabic linguistic ability.

We evaluate more than thirty Arabic and bilingual LLMs using AraLingBench and uncover consistent patterns across models. While overall performance correlates with results from general benchmarks such as ArabicMMLU, many high-scoring models achieve these results through surface-level pattern recognition or retrieval-based shortcuts rather than true linguistic understanding. These models often perform well on spelling and reading comprehension, but struggle with grammar, morphology, and syntax the very skills that underlie authentic Arabic proficiency.

By focusing on the linguistic core of Arabic understanding, AraLingBench provides a framework to diagnose where models succeed and where they fall short. It enables researchers to distinguish between models that merely reproduce fluent Arabic text and those that internalize its grammatical and morphological structure. In doing so, AraLingBench moves Arabic NLP evaluation beyond knowledge assessment toward a more complete picture of language understanding.

\textbf{Contributions.} This work makes three main contributions:  
\begin{enumerate}
\item  We introduce \textbf{AraLingBench}, a fully human-annotated benchmark dedicated to evaluating fundamental Arabic linguistic competence across grammar, morphology, spelling, reading comprehension, and syntax.  
\item  We conduct a large-scale evaluation of more than 30 Arabic and bilingual LLMs, revealing systematic weaknesses in grammatical and morphological reasoning despite strong performance on general benchmarks.  
\item  We analyze cross-benchmark relationships and show that AraLingBench captures a distinct dimension of model ability, separating genuine linguistic understanding from surface-level or retrieval-based performance.
\end{enumerate}

\section{Related Work}
\label{sec:related}

The rapid development of Arabic large language models (LLMs) has transformed the landscape of natural language processing for Arabic-speaking communities, creating an urgent need for more comprehensive and linguistically grounded evaluation frameworks. We organize this review into three main areas: (1) the evolution of Arabic language models, (2) the expansion of Arabic evaluation benchmarks, and (3) how AraLingBench addresses the persistent gap in linguistic evaluation.

\subsection{Arabic Language Models}

The development of Arabic language models has progressed through several architectural generations. The encoder-based era began with AraBERT \citep{antoun-etal-2020-arabert}, which adapted BERT's architecture for Arabic text, followed by specialized variants such as MARBERT \citep{abdul-mageed-etal-2021-arbert} for dialectal processing, ARBERT \citep{panboonyuen2025albertadvancedlocalizationbidirectional} for social media content, and CAMeLBERT \citep{inoue-etal-2021-interplay}. These models established strong baselines for Arabic NLP tasks but remained limited to discriminative applications such as classification and sequence labeling.

The transition to generative architectures marked a major turning point for Arabic NLP. AraGPT2 introduced Arabic text generation capabilities, paving the way for larger bilingual and multilingual systems. JAIS \citep{sengupta202316149v2} (13B–30B parameters) stands among the most ambitious bilingual Arabic–English efforts, while ALLaM \citep{allam2024allam} (7B–13B) emphasizes instruction tuning and task generalization. Similarly, AceGPT \citep{huang-etal-2024-acegpt} focuses on cross-lingual transfer learning built upon LLaMA-2, while models such as Hala \citep{hammoud202514008v1} leverage synthetic bilingual data for fine-tuning.

Recent contributions have diversified further. Yehia \citep{yehia2025} has consistently ranked among top-performing Arabic models across multiple benchmarks in our evaluation. Fanar \citep{fanarteam2025fanararabiccentricmultimodalgenerative}, developed by QCRI, benefits from extensive computational resources, while domain-specialized systems such as Atlas-Chat \citep{aloui2024atlas} and ArabianGPT target dialectal and regional variants, including Moroccan Darija and Gulf Arabic. Collectively, these developments illustrate the growing maturity and variety of approaches in the Arabic LLM ecosystem.

Table~\ref{tab:arabic_models_corrected_v3} summarizes major Arabic language models, outlining their architectures, scales, and training strategies. This taxonomy highlights the diversity of design choices—ranging from models trained from scratch on Arabic data to those adapted from multilingual foundations through continued pretraining or supervised fine-tuning.

\begin{table*}[t]
\centering
\caption{Comparison of major Arabic language models showing architecture types, parameter scales, and training approaches.}
\label{tab:arabic_models_corrected_v3}
\begin{threeparttable}
\resizebox{\textwidth}{!}{%
\begin{tabular}{@{}llll@{}}
\toprule
\textbf{Model} & \textbf{Size} & \textbf{Architecture} & \textbf{Training Approach} \\
\midrule
AraBERT \citep{antoun-etal-2020-arabert} & 110M / 335M & Encoder (BERT) & Pretrained from scratch \\
MARBERT \citep{abdul-mageed-etal-2021-arbert} & \(\sim\)163M & Encoder (BERT) & Pretrained from scratch \\
CAMeLBERT \citep{inoue-etal-2021-interplay} & 110M & Encoder (BERT) & Pretrained from scratch \\
JABER \citep{Ghaddar2021JABERJA} & \(\sim\)125M & Encoder (BERT) & Pretrained from scratch \\
AraGPT2 \citep{antoun-etal-2021-aragpt2} & 125M–1.5B & Decoder (GPT-2) & Pretrained from scratch \\
JAIS \citep{sengupta202316149v2}  & 13B/30B/70B & Decoder-only Transformer & Pretrained from scratch; +Chat SFT \\
ALLaM \citep{allam2024allam} & 7B / 13B / 34B / 70B & Decoder-only Transformer & Continued pretraining + SFT \\
AceGPT\citep{huang-etal-2024-acegpt} & 7B / 13B & Decoder & Continued pretraining (Llama-2) + SFT \\
Hala \citep{hammoud202514008v1} & 350M/700M/1.2B/9B & Decoder & SFT on synthetic bilingual supervision \\
Atlas-Chat \citep{aloui2024atlas}& 9B & Decoder & Dialect-focused SFT \\
Yehia \citep{yehia2025} & 7B & Decoder & Instruction tuning (SFT/DPO) \\
Fanar \citep{fanarteam2025fanararabiccentricmultimodalgenerative} & 9B & Decoder & Continued pretraining (Gemma-2-9B) + SFT \\
SUHAIL \citep{Suhail2025} & 14B & Decoder & Inst. tuning / LoRA on multilingual base \\
ArabianGPT \citep{Kouba2024ArabianGPTNA} & 1.5B & Decoder & Continued pretraining + SFT \\
Jais-Adapted \citep{jaisfamilymodelcard} & 13B & Decoder & Instruction tuning (from Llama-2) \\
\bottomrule
\end{tabular}%
} 

\begin{tablenotes}
\small 
\item \textbf{Notes:}
\begin{itemize}[leftmargin=*, label=--]
    \item \textit{Pretrained from scratch:} Trained on Arabic data from initialization.
    \item \textit{Continued pretraining:} Further trained from a multilingual base model.
    \item \textit{SFT:} Supervised fine-tuning.
    \item \textit{LoRA:} Low-rank adaptation for efficient fine-tuning.
\end{itemize}

\end{tablenotes}
\end{threeparttable}
\end{table*}

\subsection{Evaluation Benchmarks for Arabic}

Alongside advances in model development, the Arabic evaluation ecosystem has grown substantially, now encompassing over 40 distinct benchmarks \citep{alzubaidi202513430v2}. Most of these benchmarks are knowledge-intensive, designed to test factual reasoning or domain expertise rather than linguistic competence. ArabicMMLU \citep{koto-etal-2024-arabicmmlu} leads this trend, featuring over 14,000 multiple-choice questions across academic and professional subjects. EXAMS \citep{hardalov-etal-2020-exams} extends this with standardized test material, while 3LM \citep{boussaha-etal-2025-3lm} bridges Arabic, STEM, and coding domains.

Multi-task evaluation platforms have also emerged to support more holistic assessment. ORCA \citep{elmadany-etal-2023-orca} provides a unified framework for comparing models across diverse NLP tasks. AlGhafa \citep{alkhamissi2024alghafa} introduces balanced test sets that reflect both classical and modern Arabic, and BALSAM \citep{sengupta2024balsam} has evolved into a community benchmark platform featuring standardized protocols and public leaderboards.

In addition, domain-specific benchmarks now target specialized forms of reasoning and linguistic variation. For instance, ArabLegalEval \citep{hassan2024arablegaleval} tests legal reasoning, MedArabiQ \citep{daoud202503427v2} evaluates medical knowledge, and several financial datasets assess economic understanding. Cultural and dialectal benchmarks, including Palm \citep{zhang2024palm}, AraDiCE \citep{alshaikh202518442v1}, and ArabCulture \citep{sadallah-etal-2025-commonsense}, explore models’ sensitivity to regional variation and sociocultural context.

Table~\ref{tab:arabic_benchmarks_clarified} provides an overview of key Arabic benchmarks, their focus areas, and whether they directly target linguistic skills. Notably, none of the existing benchmarks explicitly evaluate core linguistic competence, which defines the unique contribution of AraLingBench.

\begin{table*}[t]
\centering
\caption{Comparison of Arabic LLM evaluation benchmarks.}
\label{tab:arabic_benchmarks_clarified}
\small
\begin{tabular}{@{}lllllll@{}}
\toprule
\textbf{Benchmark} & \textbf{Year} & \textbf{Primary Focus} & \textbf{Type} & \textbf{Source} & \textbf{Ling.} \\
\midrule
ArabicMMLU \citep{koto-etal-2024-arabicmmlu} & 2024 & Knowledge (MMLU) & MC & Native & No \\
EXAMS \citep{hardalov-etal-2020-exams} & 2020 & Knowledge (Exams) & MC & Native & No \\
AlGhafa \citep{alkhamissi2024alghafa} & 2023 & Multi-task NLP & MC & Mixed & No \\
ORCA \citep{elmadany-etal-2023-orca} & 2023 & Multi-task NLP & Mixed & Mixed & No \\
BALSAM  \citep{sengupta2024balsam} & 2025 & Platform & Mixed & Mixed & No \\
3LM \citep{boussaha-etal-2025-3lm} & 2025 & STEM + Code & Mixed & Mixed & No \\
ArabLegalEval \citep{hassan2024arablegaleval} & 2024 & Legal & Mixed & Mixed & No \\
MedArabiQ\citep{daoud202503427v2} & 2025 & Medical & Mixed & Mixed & No \\
CamelEval \citep{qian2024camelevaladvancingculturallyaligned} & 2024 & Instruction & Open & Mixed & No \\
AraDiCE \citep{alshaikh202518442v1} & 2024 & Dialectal + Cultural & Mixed & Mixed & No \\
PalmX \citep{zhang2024palm} & 2025 & Cultural & MC & Mixed & No \\
ACVA \citep{huang-etal-2024-acegpt} & 2023 & Cultural Values & MC & Mixed & No \\ \hline
\textbf{AraLingBench (Ours)} & 2025 & \textbf{Linguistic} & MC & Native & \textbf{Yes} \\
\bottomrule
\end{tabular}
\begin{tablenotes}
\footnotesize
\item \textbf{Notes:}
\begin{itemize}[leftmargin=*, label=--]
    \item \textit{MC:} Multiple-choice.
    \item \textit{Open:} Open-ended.
    \item \textit{Mixed:} Various formats.
    \item \textit{Ling.:} Focuses on core linguistic capabilities.
\end{itemize}

\end{tablenotes}
\end{table*}

\subsection{Positioning AraLingBench}

AraLingBench directly addresses the gap left by existing evaluation benchmarks by focusing on the linguistic dimension of Arabic understanding. Unlike prior benchmarks that assume linguistic competence as a prerequisite for reasoning, AraLingBench makes it the explicit target of evaluation.

The benchmark systematically assesses five fundamental categories—grammar, morphology, spelling and orthography, reading comprehension, and syntax that collectively define Arabic linguistic proficiency. Each category contains carefully constructed questions that isolate specific linguistic phenomena, allowing for precise diagnosis of model strengths and weaknesses. Because all items are authored and reviewed by human experts, AraLingBench provides a high-quality, interpretable resource for evaluating whether models have acquired genuine linguistic competence or not.

In this way, AraLingBench complements existing benchmarks while filling a critical gap: it re-centers linguistic understanding as the foundation of Arabic NLP evaluation.

\section{AraLingBench Construction}
\label{sec:method}

\subsection{Data Collection Process}

The construction of AraLingBench followed a rigorous multi-stage process designed to ensure both high quality and linguistic validity. We employed a team of five Arabic linguistics experts from the American University of Beirut, all with advanced training/degrees in Arabic grammar, morphology, and syntax. The process unfolded through four distinct phases:

\textbf{Phase 1: Question Generation.} Each expert authored original question–answer pairs across the five benchmark categories. While experts could consult reference materials such as textbooks and exam archives for inspiration, they were required to compose novel questions. This approach ensured that all items reflected authentic linguistic phenomena without reproducing existing materials.

\textbf{Phase 2: Difficulty and Diversity Filtering.} A separate group of native Arabic speakers (non-experts) reviewed the questions to assess clarity and perceived difficulty. Questions were retained if they met two conditions: (1) they were sufficiently challenging to the validation group, indicating non-trivial difficulty, and (2) they represented a diverse range of linguistic phenomena and formats.

\textbf{Phase 3: Expert Quality Control.} A senior Arabic linguist reviewed all candidate questions for accuracy, phrasing, and category alignment. Items were refined to guarantee unambiguous wording and exactly one correct answer. When a question could belong to multiple categories, it was assigned to the one most closely aligned with its primary linguistic focus.

\textbf{Phase 4: Difficulty Annotation.} Three independent annotators rated each question on a three-point scale (\{1, 2, 3\}) corresponding to Easy, Medium, and Hard. The final difficulty label was determined by majority vote.

This process produced 150 carefully validated questions evenly distributed across the five linguistic categories (30 per category), forming a compact, fully human-authored multiple choice benchmark explicitly focused on core Arabic linguistic skills for LLM evaluation.

\subsection{Benchmark Statistics}

\begin{figure}[H]
    \centering
    \includegraphics[width=1\linewidth]{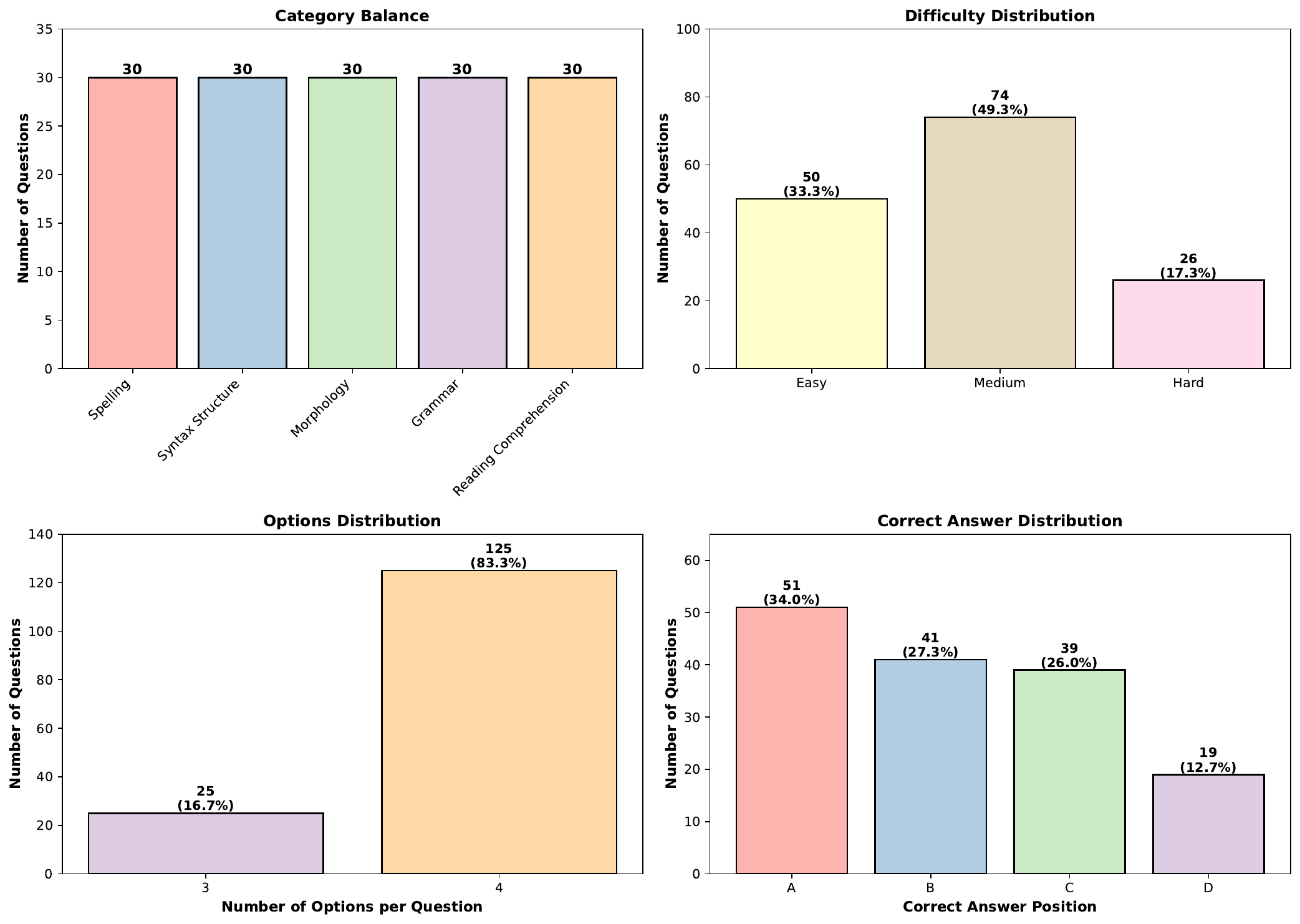}
    \caption{\textbf{Overview of AraLingBench.} Category balance, difficulty distribution, question formats, and answer position frequencies. The benchmark maintains balanced coverage across linguistic categories and difficulty levels.}
    \label{fig:dataset_stats}
\end{figure}

Figure~\ref{fig:dataset_stats} presents key statistics for AraLingBench. The dataset maintains perfect category balance, with 30 questions per category. The difficulty distribution comprises 50 Easy (33.3\%), 74 Medium (49.3\%), and 26 Hard (17.3\%) items—an intentional skew toward medium difficulty to maximize discriminative power while preserving a range of challenge levels.

Regarding format, 125 questions (83.3\%) use four-choice multiple-choice format, while 25 (16.7\%) use three options. The correct answer distribution shows moderate variation across positions (A: 34.0\%, B: 27.3\%, C: 26.0\%, D: 12.7\%), reflecting natural variation in item design without systematic positional bias.

\section{Experimental Evaluation}
\label{sec:experiments}

We evaluate over 30 Arabic and bilingual large language models on AraLingBench to assess their linguistic competence. Our analysis is structured around four key research questions, each probing a distinct dimension of model performance and benchmark validity.

\subsection{Evaluation Setup}

\textbf{Model Selection.} We evaluated 35 models drawn from the Open Arabic LLM Leaderboard, spanning sizes from 350M to 70B parameters. The set includes Arabic-specific models (Hala, Fanar, Yehia), bilingual systems (JAIS, ALLaM), and multilingual base models adapted for Arabic (Qwen2.5, Phi-4). This diversity enables comprehensive comparison across architectures, training strategies, and parameter scales.

\textbf{Evaluation Protocol.} All models were tested using zero-shot prompting in a standardized multiple-choice format. Each question was presented fully in Arabic with answer options labeled A–D. Models produced a single-letter response, which was automatically matched to the gold label. No few-shot examples or chain-of-thought prompting were used, ensuring a uniform and fair evaluation setting.

\textbf{Metrics.} We report accuracy (\%) per linguistic category and compute overall average accuracy across all five categories. Accuracy provides a direct and interpretable measure of linguistic competence and allows straightforward comparison with existing Arabic benchmarks.

\subsection{RQ1: Do Models Exhibit Balanced Linguistic Competence?}

\textbf{Motivation.} True Arabic linguistic competence requires mastery across spelling, syntax, morphology, grammar, and reading comprehension. We analyze whether models achieve balanced proficiency or exhibit specialized strengths and weaknesses.

\textbf{Results.} Table~\ref{tab:main_results} reports per-category performance for all evaluated models. Several trends emerge:

\begin{enumerate}
    \item \textbf{Performance Stratification:} Models cluster into three clear tiers. Top performers (Yehia-7B, ALLaM-7B) achieve 72–74\% average accuracy. Mid-tier models (Fanar, Qwen2.5-14B variants) reach 55–62\%, while smaller or less specialized models remain below 50\%.
    
    \item \textbf{Category Difficulty:} Spelling and Reading Comprehension are the easiest (median $\sim$58–60\%), while Syntax proves most challenging (median $\sim$48\%). This aligns with linguistic theory—surface-level orthographic patterns are easier to internalize than deep syntactic structures.
    
    \item \textbf{Unbalanced Competence:} Even the strongest models show substantial variance across categories. Yehia-7B attains 86.7\% on Spelling but only 53.3\% on Syntax—a 33-point gap—indicating asymmetric skill development.
    
    \item \textbf{Morphology Challenge:} Despite Arabic’s morphological richness, models consistently underperform in this category (median $\sim$60\%). High performers like Yehia reach only 80\%, suggesting persistent limitations in representing derivational and inflectional patterns.
\end{enumerate}

\begin{figure}[H]
    \centering
    \includegraphics[width=0.75\linewidth]{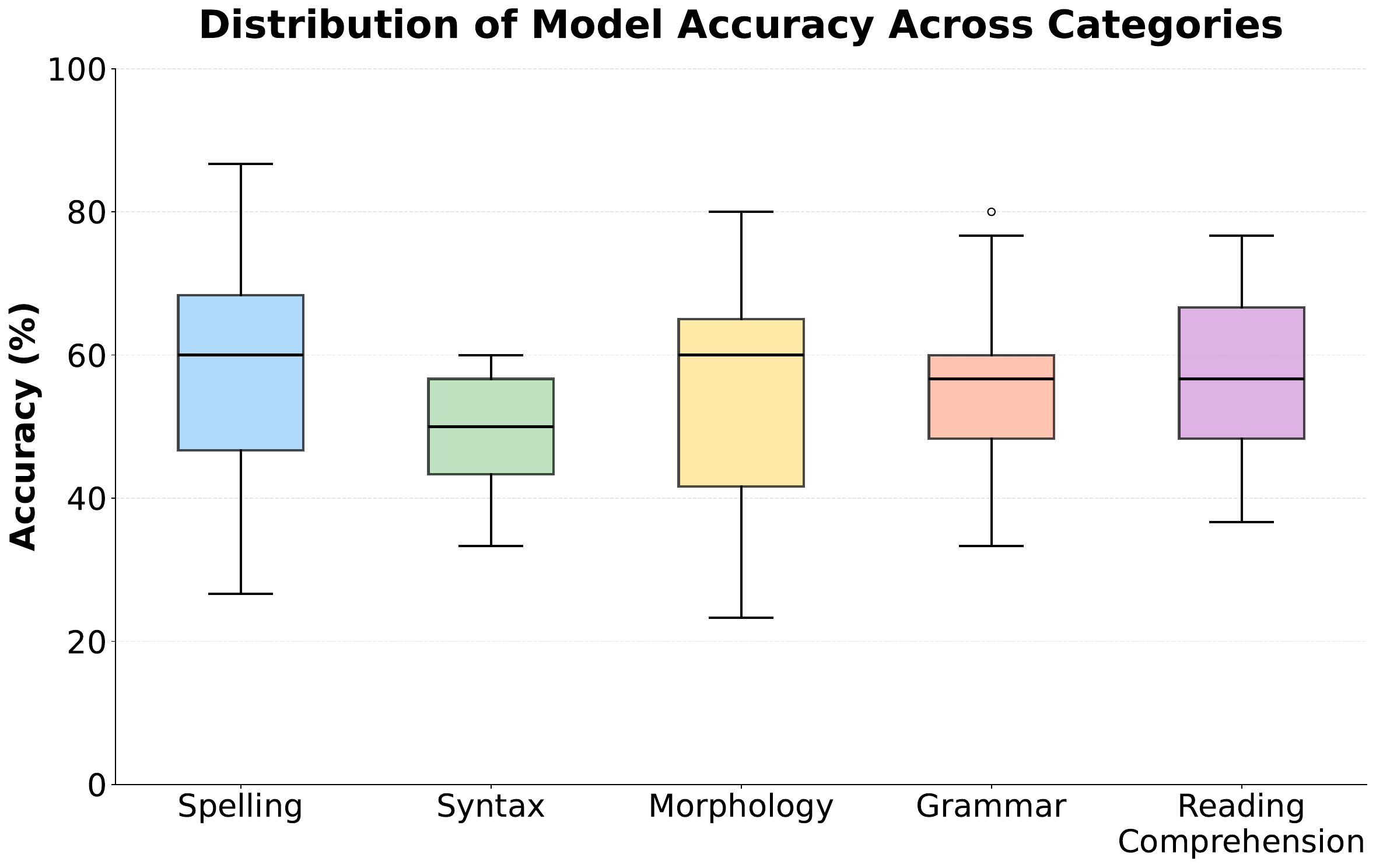}
    \caption{\textbf{Category-level accuracy distribution.}  Models perform best on Spelling and Reading Comprehension, with Syntax remaining the most difficult category.}
    \label{fig:category_boxplot}
\end{figure}

Figure~\ref{fig:category_boxplot} visualizes category-level performance variability. Wide interquartile ranges (15–20 percentage points) indicate substantial heterogeneity in how architectures and training regimes affect each linguistic dimension.

\textbf{Interpretation.} Overall, current Arabic LLMs lack balanced linguistic competence. Most models prioritize surface-level tasks (spelling, lexical retrieval) over structural understanding (syntax, morphology). This imbalance likely reflects training corpus composition, where correct orthography and frequent lexical patterns are abundant but explicit grammatical or morphological signals are sparse.

\begin{table*}[t!]
\centering
\small
\setlength{\tabcolsep}{6pt}
\resizebox{\textwidth}{!}{
\begin{tabular}{lrrrrrr}
\toprule
\textbf{Model} & \textbf{Spelling} & \textbf{Syntax} & \textbf{Morphology} & \textbf{Grammar} & \textbf{Reading Comp.} & \textbf{Average} \\
\midrule
Yehia-7B-preview & 86.7 & 53.3 & 80.0 & 80.0 & 70.0 & \textcolor{red} {74.0} \\
ALLaM-7B-Instruct-preview & 86.7 & 60.0 & 73.3 & 73.3 & 76.7 & \textcolor{red}{74.0} \\
Yehia-7B-Reasoning-preview & 80.0 & 50.0 & 80.0 & 76.7 & 73.3 & 72.0 \\
Yehia-7B-DPO-Reasoning-preview & 80.0 & 50.0 & 80.0 & 76.7 & 73.3 & 72.0 \\
Yehia-7B-SFT-Reasoning-preview & 76.7 & 36.7 & 66.7 & 76.7 & 73.3 & 66.0 \\
tempmotacilla-cinerea-0308 & 63.3 & 60.0 & 60.0 & 60.0 & 70.0 & 62.7 \\
Qwen2.5-Lumen-14B & 70.0 & 56.7 & 63.3 & 60.0 & 60.0 & 62.0 \\
Saka-14B & 66.7 & 56.7 & 63.3 & 60.0 & 60.0 & 61.3 \\
SUHAIL-14B-preview & 60.0 & 60.0 & 70.0 & 63.3 & 53.3 & 61.3 \\
lambda-qwen2.5-14b-dpo-test & 70.0 & 60.0 & 60.0 & 60.0 & 56.7 & 61.3 \\
Qwen2.5-14B & 60.0 & 46.7 & 60.0 & 70.0 & 66.7 & 60.7 \\
Qwen2.5-14B-Gutenberg-1e-Delta & 70.0 & 56.7 & 60.0 & 60.0 & 56.7 & 60.7 \\
Rombos-LLM-V2.6-Qwen-14b & 70.0 & 60.0 & 60.0 & 56.7 & 56.7 & 60.7 \\
Fanar-1-9B-Instruct & 60.0 & 43.3 & 73.3 & 63.3 & 60.0 & 60.0 \\
Qwen2.5-14B-Instruct & 66.7 & 56.7 & 60.0 & 56.7 & 53.3 & 58.7 \\
Qwen3-8B-Base & 60.0 & 60.0 & 60.0 & 60.0 & 43.3 & 56.7 \\
Hala-9B & 63.3 & 40.0 & 63.3 & 46.7 & 60.0 & 54.7 \\
emirati-14b-v2 & 63.3 & 46.7 & 60.0 & 56.7 & 46.7 & 54.7 \\
Qwen2.5-7B-Instruct-abliterated-v2 & 53.3 & 53.3 & 43.3 & 56.7 & 60.0 & 53.3 \\
SILMA-9B-Instruct-v1.0 & 53.3 & 53.3 & 66.7 & 56.7 & 36.7 & 53.3 \\
T.E-8.1 & 43.3 & 43.3 & 60.0 & 50.0 & 66.7 & 52.7 \\
Marco-LLM-AR-V2 & 53.3 & 46.7 & 66.7 & 56.7 & 36.7 & 52.0 \\
recoilme-gemma-2-9B-v0.4 & 56.7 & 40.0 & 63.3 & 53.3 & 46.7 & 52.0 \\
Qwen2.5-7B-Instruct & 50.0 & 50.0 & 40.0 & 50.0 & 66.7 & 51.3 \\
Qwen2.5-7B-Instruct-Uncensored & 46.7 & 40.0 & 46.7 & 50.0 & 66.7 & 50.0 \\
Josiefied-Qwen2.5-7B & 43.3 & 50.0 & 40.0 & 46.7 & 60.0 & 48.0 \\
Qwen2.5-7B-Instruct-abliterated & 46.7 & 40.0 & 40.0 & 46.7 & 66.7 & 48.0 \\
SauerkrautLM-Nemo-12b-Instruct & 46.7 & 40.0 & 46.7 & 43.3 & 56.7 & 46.7 \\
Phi-4-mini-instruct & 46.7 & 43.3 & 53.3 & 43.3 & 43.3 & 46.0 \\
Qwen2-7B-Instruct & 50.0 & 46.7 & 36.7 & 43.3 & 50.0 & 45.3 \\
Marco-LLM-AR-V4 & 53.3 & 33.3 & 40.0 & 50.0 & 36.7 & 42.7 \\
Qwen2.5-3B-Instruct & 36.7 & 50.0 & 26.7 & 50.0 & 36.7 & 40.0 \\
Hala-1.2B & 26.7 & 43.3 & 30.0 & 36.7 & 56.7 & 38.7 \\
Hala-350M & 36.7 & 43.3 & 30.0 & 46.7 & 36.7 & 38.7 \\
Hala-700M & 43.3 & 36.7 & 23.3 & 33.3 & 53.3 & 38.0 \\
\bottomrule
\end{tabular}
}
\caption{\textbf{Model performance across AraLingBench categories.} Top-performing models reach 72–74\% accuracy but show large intra-category variance, with Syntax persistently posing the greatest challenge, top models are highlited in \textcolor{red}{red}.}
\label{tab:main_results}
\end{table*}

\subsection{RQ2: How Do Linguistic Skills Correlate?}

\textbf{Motivation.} Understanding how linguistic categories relate helps determine whether Arabic competence develops holistically or through independent skill acquisition. Strong correlations suggest shared underlying representations, whereas weak ones imply category-specific learning.

\begin{figure}[h!]
    \centering
    \includegraphics[width=0.75\linewidth]{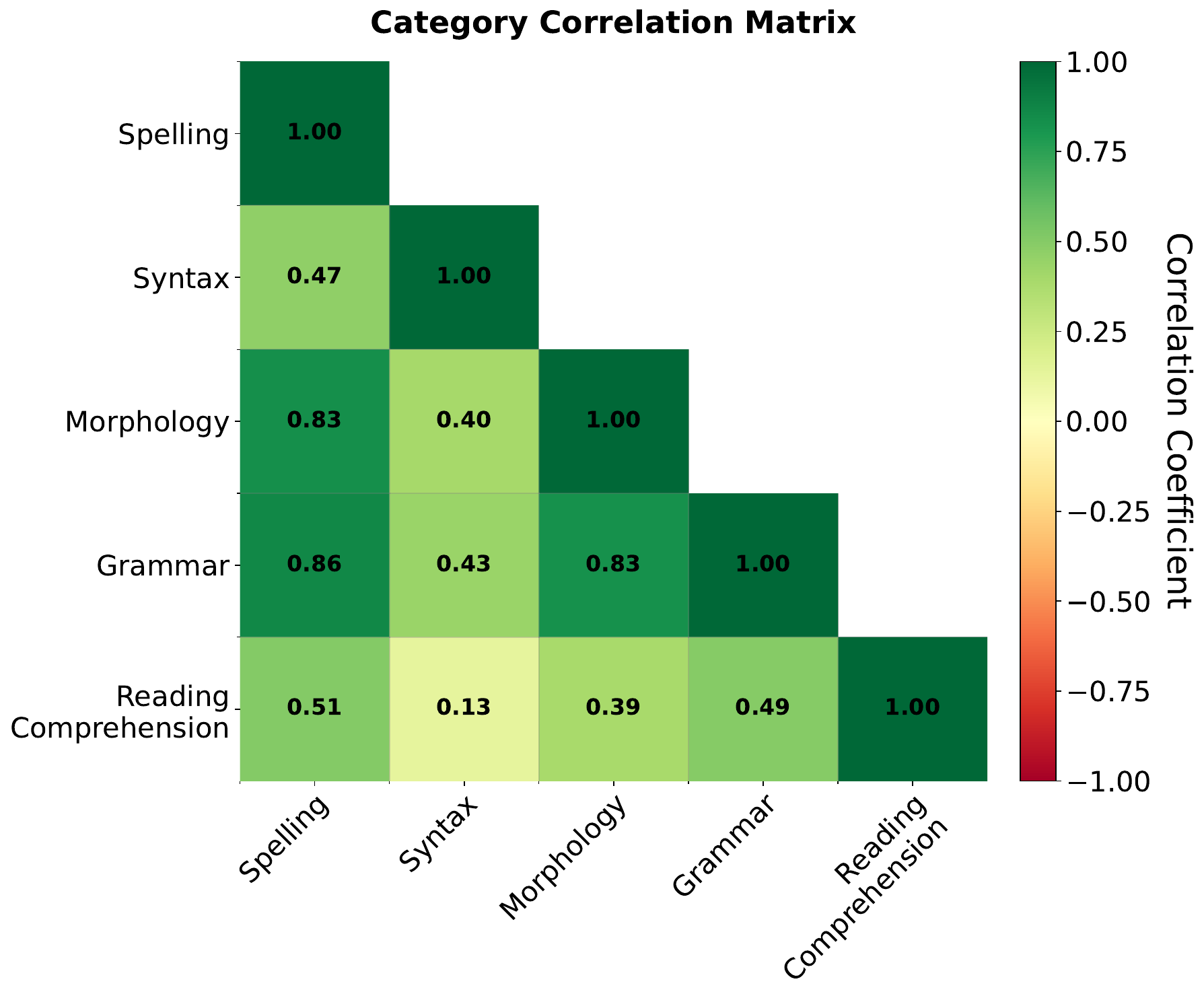}
    \caption{\textbf{Inter-category correlations.} Grammar and Morphology show the strongest relationship ($r = 0.80$), while Syntax remains comparatively independent, suggesting distinct representational mechanisms.}
    \label{fig:category_correlation}
\end{figure}

\textbf{Results.} Figure~\ref{fig:category_correlation} presents the correlation matrix across the five linguistic categories. Several clear relationships emerge:

\begin{enumerate}
    \item \textbf{Grammar–Morphology:} High correlation ($r = 0.83$) indicates that these two skills develop in tandem. Models capable of accurate morphological parsing tend to perform well in grammatical reasoning, reflecting their shared dependence on word structure and inflection.
    
    \item \textbf{Spelling–Grammar:} A similarly high correlation ($r = 0.86$) underscores the link between orthographic accuracy and grammatical knowledge. Correct spelling often depends on syntactic case or agreement, showing that orthography and grammar are mutually reinforcing.
    
    \item \textbf{Spelling–Reading Comprehension:} A moderate correlation ($r \approx 0.51$) suggests that lexical precision supports text comprehension, though understanding sentences extends beyond word-level recognition.
    
    \item \textbf{Syntax Independence:} Syntax exhibits the weakest correlations with other categories ($r \approx 0.13-0.40$), suggesting that syntactic reasoning relies on distinct mechanisms. This aligns with linguistic theory: syntactic competence requires compositional and hierarchical understanding that does not emerge directly from surface-level distributional patterns.
\end{enumerate}

\textbf{Interpretation.} These correlations indicate that Arabic linguistic competence in LLMs emerges through partially overlapping clusters of skills rather than as a single unified construct. Morphology and grammar form a tightly linked subsystem, while syntax stands apart as a specialized ability that may require targeted inductive biases—such as hierarchical attention or structural modeling—beyond those learned from general pretraining.

\subsection{RQ3: Does General Benchmark Performance Predict Linguistic Competence?}

\textbf{Motivation.} Arabic LLMs are often ranked by performance on general benchmarks such as ArabicMMLU or EXAMS. We examine whether such rankings reflect genuine linguistic understanding or merely general reasoning ability.

\textbf{Results.} Table~\ref{tab:crossbench_plus_ours} compares model scores across AraLingBench and seven established Arabic benchmarks. Several trends are evident:

\begin{enumerate}
    \item \textbf{ Predictive Alignment:} High performance on ArabicMMLU does not reliably translate to strong results on AraLingBench. For instance, Hala-9B attains 65.6\% on ArabicMMLU but only 54.7\% on AraLingBench, showing that factual or reasoning benchmarks may not capture linguistic depth, in fact, the top performers on ArabicMMLU are not necessarily the top performers on AraLingBench.
    
    \item \textbf{Synthetic Training Effects:} Models heavily tuned on synthetic instruction data (e.g., Hala family) achieve high scores on knowledge benchmarks but underperform on linguistic ones, suggesting overreliance on memorized or templated patterns.
    
    \item \textbf{Instruction-Tuning Benefits:} Models fine-tuned with real instruction data (Yehia, ALLaM) demonstrate stronger alignment between general and linguistic competence, indicating that linguistically grounded supervision improves generalization.
    
    \item \textbf{Domain Mismatch:} Some models excel on specialized domains (e.g., legal or medical QA) yet perform poorly on basic linguistic tasks, reflecting narrow domain optimization without underlying language mastery.
\end{enumerate}

\begin{table*}[t!]
\centering
\caption{\textbf{Cross-benchmark comparison.} Performance of major Arabic LLMs across eight benchmarks shows limited predictive power between knowledge-based and linguistic evaluations.}
\label{tab:crossbench_plus_ours}
\setlength{\tabcolsep}{4pt}
\renewcommand{\arraystretch}{1.08}
\footnotesize
\begin{threeparttable}
\resizebox{\textwidth}{!}{%
\begin{tabular}{@{}lrrrrrrrr@{}}
\toprule
\textbf{Model} & \textbf{AlGhafa} & \textbf{ArabicMMLU} & \textbf{EXAMS} & \textbf{MadinahQA} & \textbf{AraTrust} & \textbf{ALRAGE} & \textbf{ArbMMLU-HT} & \textbf{AraLingBench} \\
\midrule
Navid-AI/Yehia-7B-preview & 70.8 & 64.9 & 52.1 & 54.4 & 87.5 & 76.6 & 53.4 & 74.0 \\
ALLaM-7B-Instruct-preview & 69.5 & 64.9 & 51.6 & 54.2 & 86.9 & 76.8 & 52.8 & 74.0 \\
Yehia-7B-Reasoning-preview & 75.2 & 66.3 & 52.7 & 55.0 & 80.8 & 73.3 & 55.3 & 72.0 \\
Hala-9B & 78.3 & 65.6 & 53.8 & 70.4 & 89.6 & -- & 61.4 & 54.7 \\
Fanar-1-9B-Instruct & 76.4 & 65.8 & 52.7 & 73.4 & 88.3 & 77.0 & 58.6 & 60.0 \\
Qwen2.5-14B-Instruct & 72.3 & 60.0 & 53.6 & 35.6 & 86.1 & 78.9 & 55.7 & 58.7 \\
Qwen2.5-7B-Instruct & 65.6 & 52.3 & 39.7 & 62.7 & 80.7 & 77.4 & 40.3 & 51.3 \\
Hala-1.2B & 59.2 & 48.6 & 43.4 & 41.6 & 71.7 & -- & 44.2 & 38.7 \\
Hala-700M & 55.5 & 45.9 & 40.6 & 34.7 & 65.2 & -- & 39.4 & 38.0 \\
Hala-350M & 51.4 & 41.2 & 36.9 & 34.5 & 52.1 & -- & 35.4 & 38.7 \\
\bottomrule
\end{tabular}%
}
\end{threeparttable}
\end{table*}

\begin{figure}[t]
    \centering
    \includegraphics[width=0.75\linewidth]{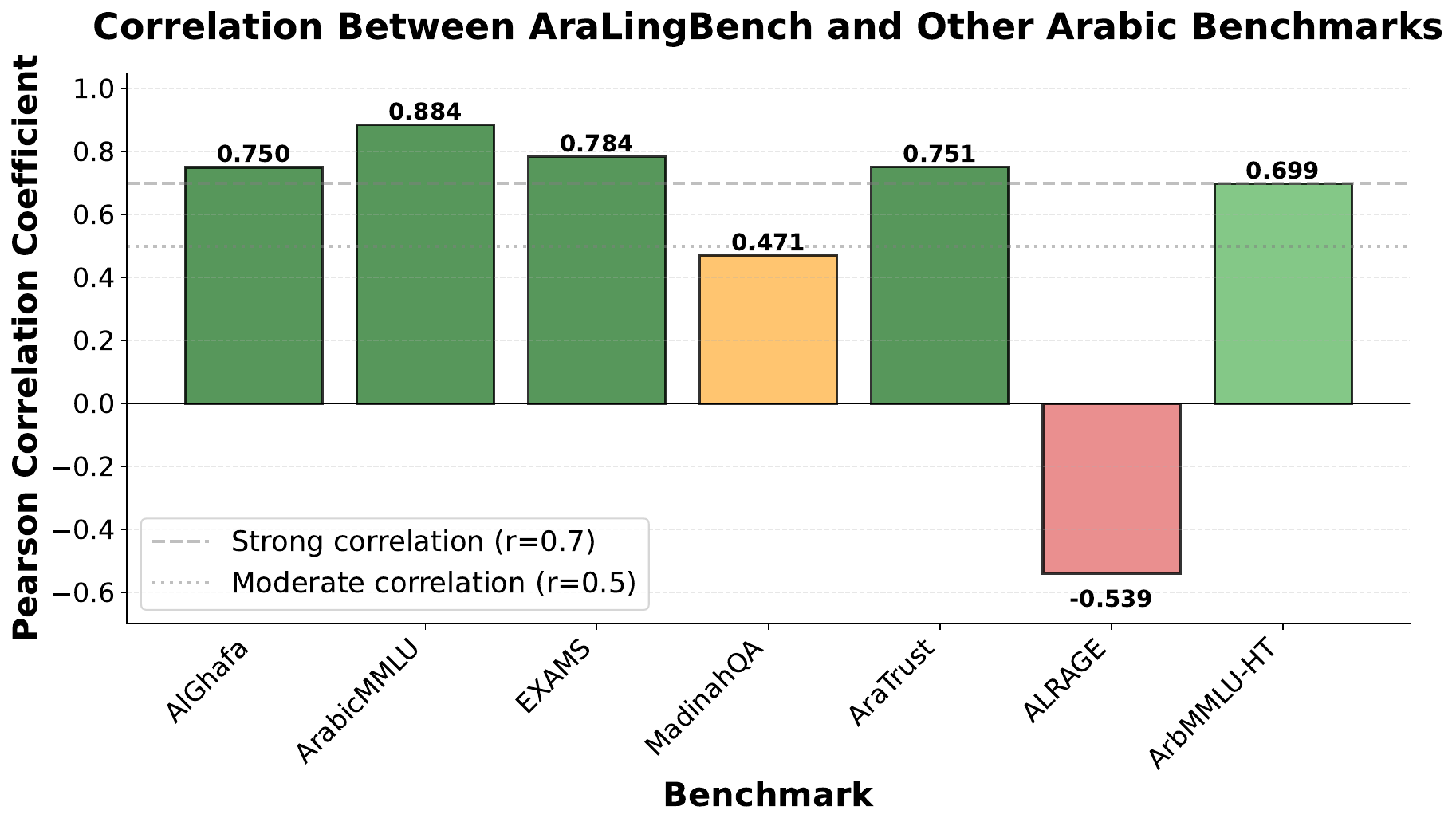}
    \caption{\textbf{Cross-benchmark correlations.} Pearson coefficients between AraLingBench and seven major Arabic benchmarks reveal strong alignment with language understanding tasks but weak or negative correlation with retrieval-augmented systems.}
    \label{fig:crossbench_correlation}
\end{figure}

\textbf{Interpretation.} Figure~\ref{fig:crossbench_correlation} highlights nuanced relationships across benchmarks. AraLingBench correlates strongly with language understanding benchmarks such as ArabicMMLU ($r=0.884$) \textit{though the correlation varies across individual MMLU categories}, EXAMS ($r=0.784$), and AraTrust ($r=0.751$), confirming that linguistic competence underpins general reasoning performance. However, negative correlation with retrieval-augmented evaluation (ALRAGE, $r=-0.539$) reveals that reliance on retrieval can substitute for genuine understanding. These findings indicate that while linguistic ability transfers positively to most NLP tasks, some optimization pathways especially synthetic or retrieval-heavy ones decouple apparent performance from true linguistic competence.

\subsection{RQ4: Does Question Difficulty Align with Model Performance?}

\textbf{Motivation.} We test whether human-annotated difficulty levels correspond to model performance patterns and whether difficulty scales consistently across models and linguistic categories.

\begin{figure}[H]
    \centering
    \includegraphics[width=0.65\linewidth]{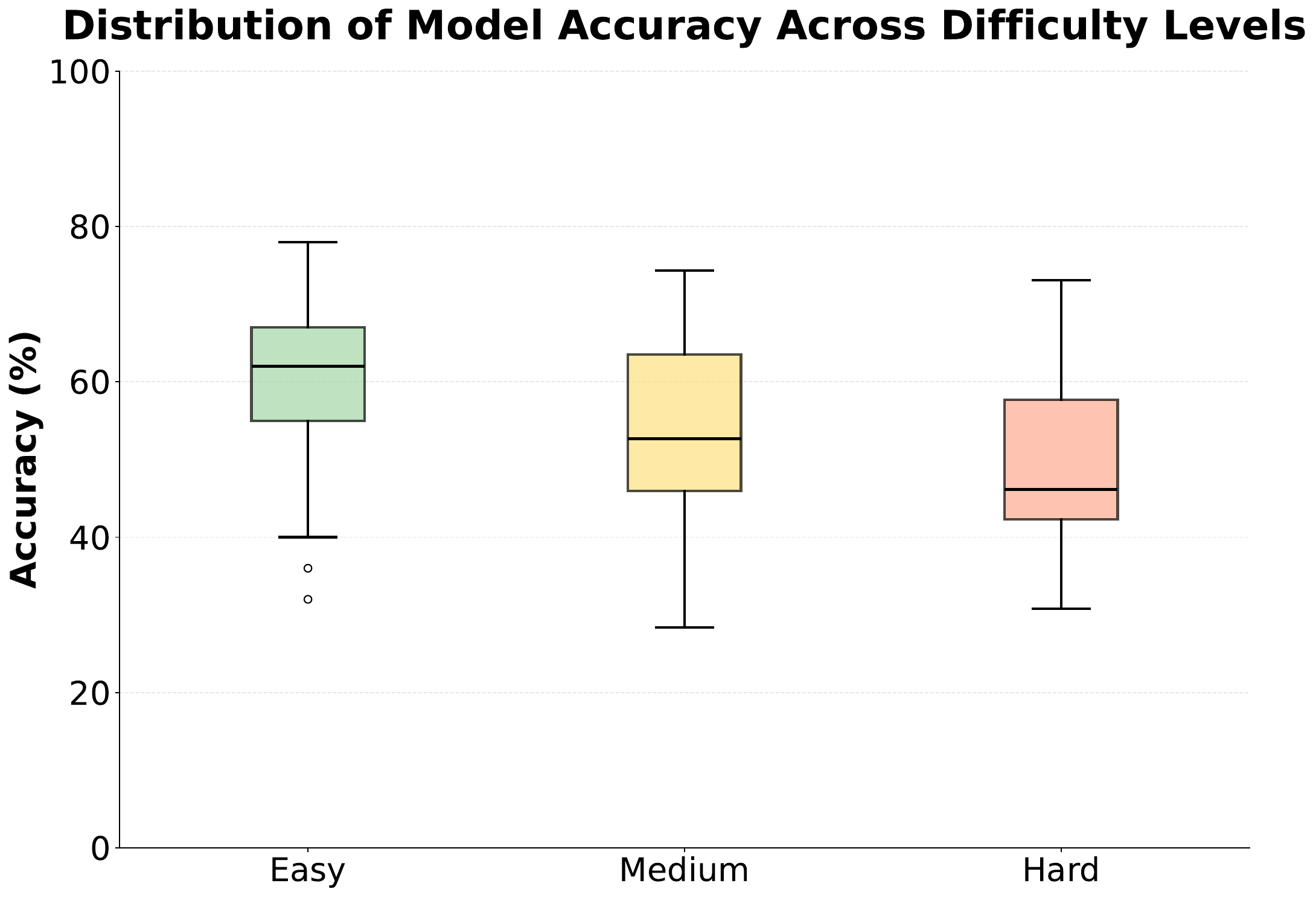}
    \caption{\textbf{Performance by difficulty level.} Model accuracy does not decrease monotonically with annotated difficulty; Hard questions occasionally yield higher accuracy than Medium ones.}
    \label{fig:difficulty_boxplot}
\end{figure}

\textbf{Results.} Figure~\ref{fig:difficulty_boxplot} shows accuracy distributions across Easy, Medium, and Hard questions.

\begin{enumerate}
    \item \textbf{Non-Monotonic Scaling:} Performance does not degrade consistently with difficulty. Easy questions yield a median accuracy of 58\%, Medium 50\%, but Hard questions recover to 54\%.
    
    \item \textbf{Model-Dependent Difficulty:} Certain models (e.g., Qwen3-8B-Base) perform better on Hard items (73.1\%) than Medium ones (50.0\%), indicating a mismatch between human and model difficulty perception.
    
    \item \textbf{Category Interactions:} Difficulty effects vary by category. Hard Morphology questions can be easier for models than Medium ones, while Syntax maintains a consistent downward trend with difficulty.
    
    \item \textbf{Stable High-Performer Trends:} Leading models (Yehia, ALLaM) exhibit only modest degradation across levels (76\% → 69\%), suggesting robust linguistic generalization.
\end{enumerate}

\begin{figure}[H]
    \centering
    \includegraphics[width=0.95\linewidth]{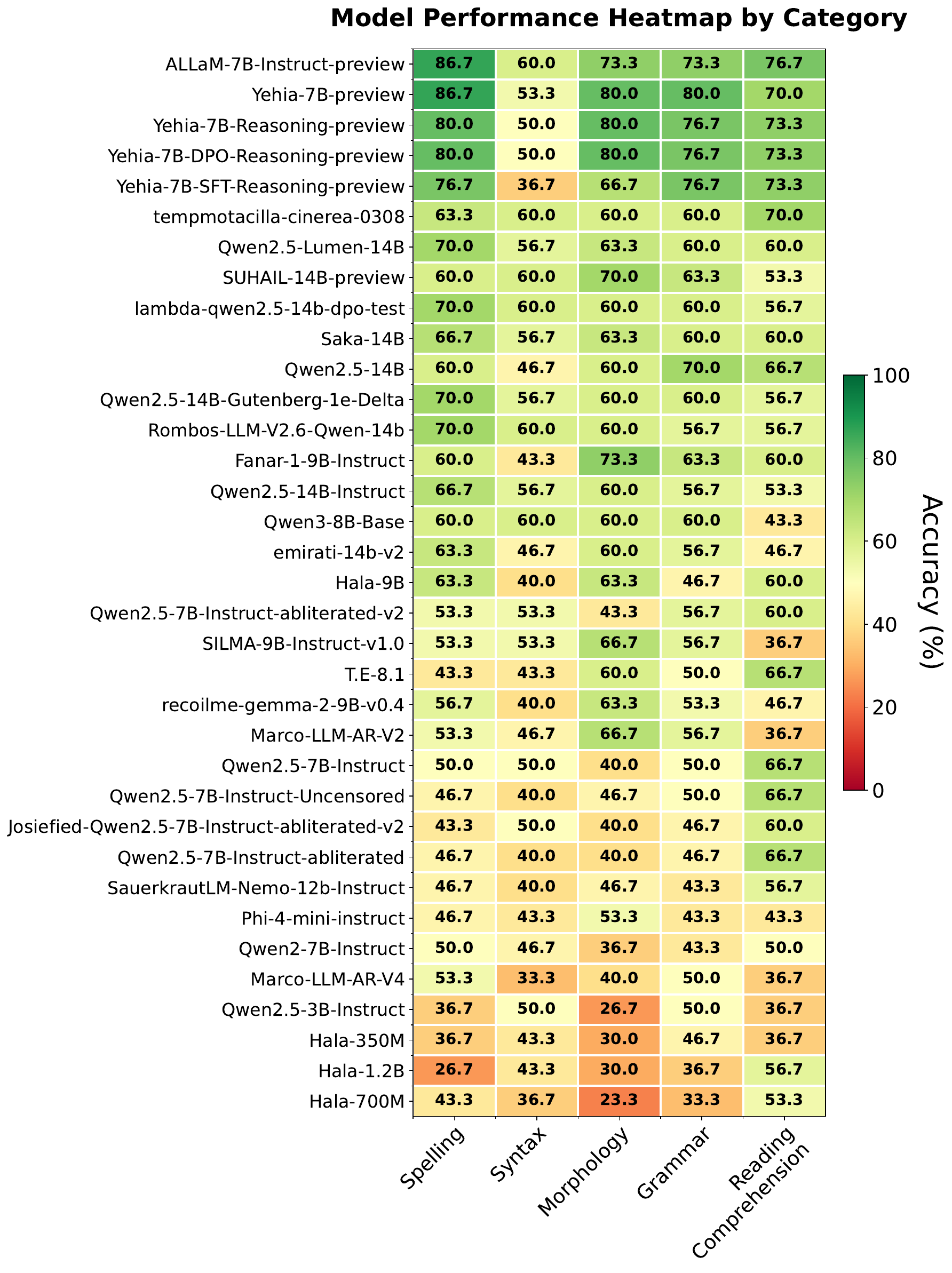}
    \caption{Model performance heatmap across the five AraLingBench linguistic categories. Accuracy values are shown for 35 evaluated models, sorted by weighted average performance. Color intensity ranges from red (low) through yellow (moderate) to green (high).}
\label{fig:category_heatmap}

    \label{fig:category_heatmap}
    \end{figure}

\begin{figure}[H]
    \centering
    \includegraphics[width=0.9\linewidth]{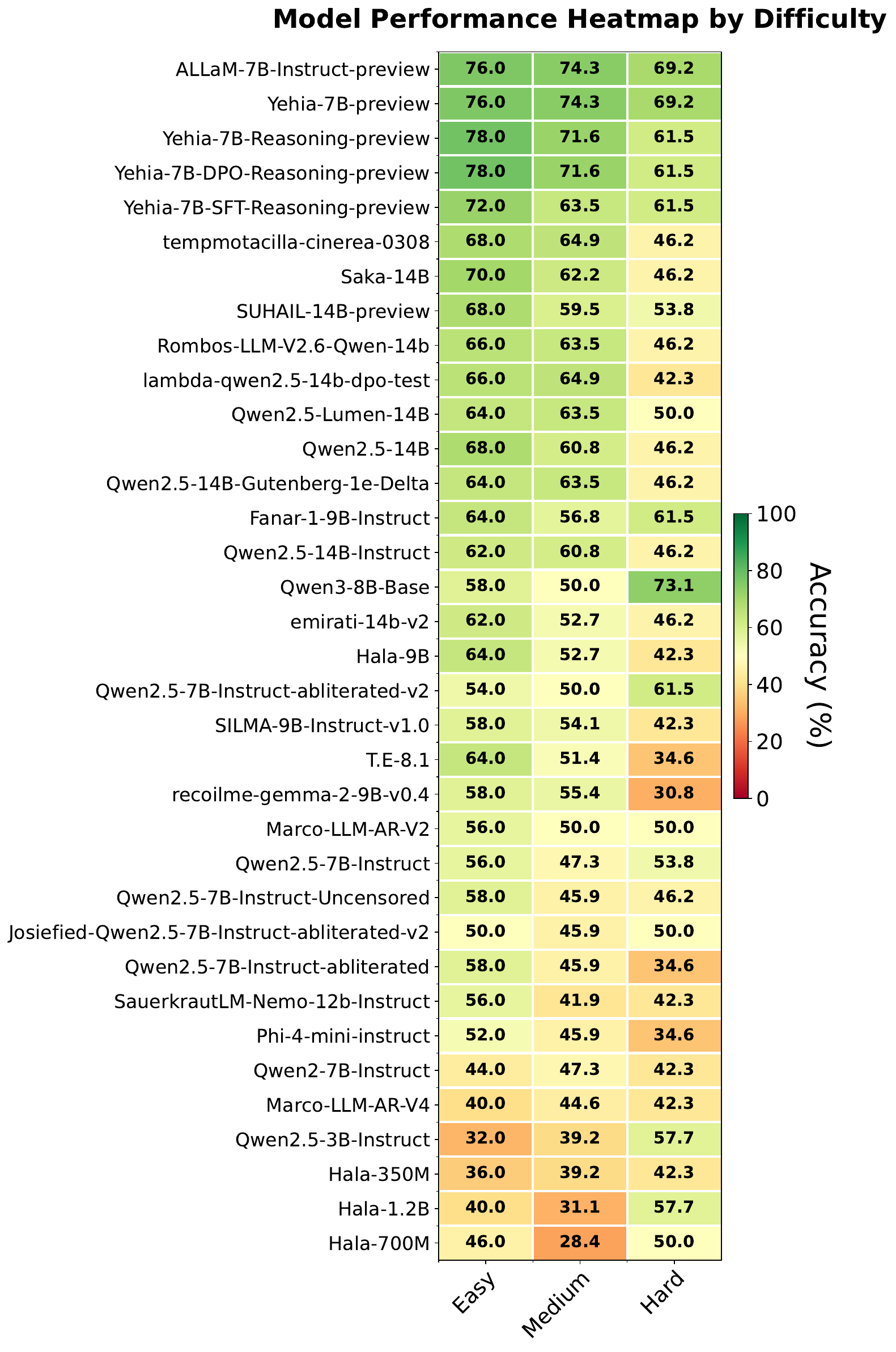}
    \caption{Model performance heatmap across AraLingBench difficulty levels (Easy, Medium, Hard) for 35 evaluated models, sorted by weighted average performance. Color intensity ranges from red (low accuracy) through yellow (moderate) to green (high).}
    \label{fig:difficulty_heatmap}
\end{figure}

\subsection{Detailed Performance Visualization}

We present two heatmaps to highlight model behavior across categories and difficulty levels.

\textbf{Category-Level Patterns.} Figure \ref{fig:category_heatmap} shows model accuracy across the five linguistic categories. Top models (Yehia-7B-preview, ALLaM-7B-Instruct-preview, 74.0\%) perform consistently well, especially in Spelling (up to 86.7\%), suggesting effective instruction tuning. Mid-range models (50--62\%) show uneven performance, with strengths in one area offsetting weaknesses in another. For example, SILMA-9B-Instruct-v1.0 scores 66.7\% in Morphology despite a 53.3\% average. Variability is highest in Reading Comprehension and Grammar, reflecting diverse alignment strategies. Models below 50\% struggle most with Morphology and Syntax (some below 30\%), showing the challenge of Arabic’s complex structure. Even top models hit a ceiling in Syntax (best: 60.0\%).

\begin{figure}[H]
    \centering
    \includegraphics[width=0.6\linewidth]{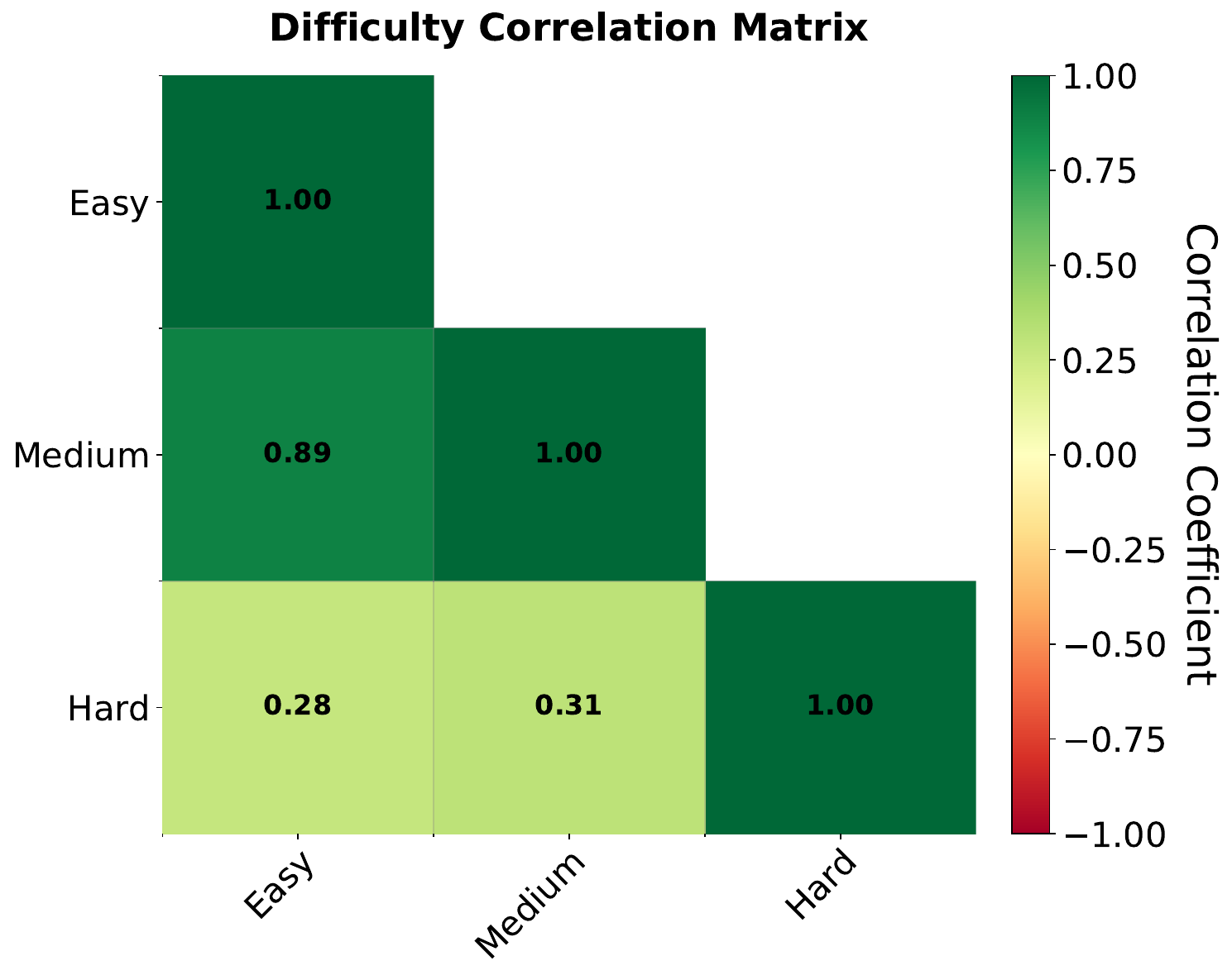}
    \caption{\textbf{Difficulty-level correlations.} Strong positive relationships ($r > 0.65$) indicate consistent model ranking despite non-monotonic accuracy patterns.}
    \label{fig:difficulty_correlation}
\end{figure}

\textbf{Difficulty-Level Patterns.} Figure \ref{fig:difficulty_correlation} presents performance across Easy, Medium, and Hard questions. Top models show stable trends with only slight drops (about 7 points from Easy to Hard). Mid-range models show irregularities: Qwen3-8B-Base scores 58.0\% (Easy), 50.0\% (Medium), and 73.1\% (Hard), indicating non-linear difficulty perception. Hard items may depend on memorized patterns or domain-specific knowledge, while Medium ones require reasoning less present in pretraining. Low-performing models show unpredictable scaling, sometimes scoring near chance on Easy but higher on Hard. This suggests a mismatch between human and model-perceived difficulty.

Overall, models acquire surface skills more reliably than deep structural understanding, and human difficulty ratings do not always predict actual model challenge.

\textbf{Interpretation.} The non-monotonic scaling pattern reveals that model difficulty diverges from human-perceived complexity. Hard questions may feature constructions or vocabulary frequent in pretraining corpora, making them easier for models, while Medium questions often require integrative reasoning that transformer models handle less effectively. This underscores the need to calibrate benchmark difficulty using both human annotation and pilot testing on representative models.

\clearpage

\subsection{Summary of Experimental Findings}

Our evaluation yields four main insights:

\begin{enumerate}
    \item Arabic LLMs display highly uneven linguistic competence, excelling in surface-level abilities (spelling, comprehension) but struggling with deeper structural understanding (syntax, morphology).
    
    \item Linguistic skills correlate moderately but not uniformly: grammar and morphology form a tightly coupled subsystem, while syntax remains largely independent.
    
    \item General benchmark success does not guarantee linguistic competence. Although overall correlations are strong ($r > 0.75$), certain training regimes especially retrieval-heavy or synthetic setups inflate benchmark scores without improving true linguistic understanding.
    
    \item Human-assigned difficulty labels only partially align with model performance, highlighting the need to jointly consider cognitive and data-driven measures of challenge.
\end{enumerate}

Taken together, these findings establish AraLingBench as a crucial complement to existing Arabic evaluation suites. It isolates fundamental linguistic understanding, exposing competence gaps that remain invisible in knowledge-oriented benchmarks.

\section{Conclusion}
\label{sec:conclusion}

We introduced \textit{AraLingBench}, a fully human annotated benchmark designed to evaluate the fundamental linguistic competence of Arabic large language models. By focusing on five core categories grammar, morphology, spelling, reading comprehension, and syntax AraLingBench isolates the linguistic foundations of Arabic understanding that current knowledge-based benchmarks overlook.  

Our large-scale evaluation of more than 30 models reveals that high performance on general benchmarks does not necessarily indicate genuine language understanding, as many models continue to struggle with grammatical and morphological reasoning. AraLingBench offers a complementary diagnostic perspective, enabling researchers to distinguish between superficial fluency and true linguistic competence.  

We release the benchmark to support the development of Arabic LLMs that not only generate fluent text but also demonstrate authentic mastery of the language’s structure and logic.

\bibliography{main}
\bibliographystyle{tmlr}

\appendix

\end{document}

%% file: math_commands.tex
\usepackage{amsmath,amsfonts,bm}

\def\eqref#1{equation~\ref{#1}}

\def\1{\bm{1}}

\DeclareMathAlphabet{\mathsfit}{\encodingdefault}{\sfdefault}{m}{sl}
\SetMathAlphabet{\mathsfit}{bold}{\encodingdefault}{\sfdefault}{bx}{n}